\documentclass{article}
\usepackage[T1]{fontenc} 
\usepackage[utf8]{inputenc} 
\usepackage{ismir,amsmath,cite,url}
\usepackage{graphicx}
\usepackage{color}
\usepackage{multirow}

\usepackage{lineno}

\title{Camera-Based Piano Sheet Music Identification}

\twoauthors
  {Daniel Yang} {Harvey Mudd College \\ {\tt dhyang@g.hmc.edu}}
  {TJ Tsai} {Harvey Mudd College \\ {\tt ttsai@g.hmc.edu}}

\begin{document}

\maketitle
\begin{abstract}
This paper presents a method for large-scale retrieval of piano sheet music images.  Our work differs from previous studies on sheet music retrieval in two ways.  First, we investigate the problem at a much larger scale than previous studies, using all solo piano sheet music images in the entire IMSLP dataset as a searchable database.  Second, we use cell phone images of sheet music as our input queries, which lends itself to a practical, user-facing application.  We show that a previously proposed fingerprinting method for sheet music retrieval is far too slow for a real-time application, and we diagnose its shortcomings.  We propose a novel hashing scheme called dynamic n-gram fingerprinting that significantly reduces runtime while simultaneously boosting retrieval accuracy.  In experiments on IMSLP data, our proposed method achieves a mean reciprocal rank of $0.85$ and an average runtime of $0.98$ seconds per query.
\end{abstract}
\section{Introduction}
\label{sec:introduction}

Imagine the following scenario.  A musician is sitting down in front of a piano learning a new piece of music.  She pulls out her cell phone, takes a picture of the physical page of sheet music sitting in front of her, and is immediately able to access Youtube videos of performances of that piece and alternate editions of the sheet music.  In this paper, we present a method to solve the main technical challenge of identifying the page of music.  This is the camera-based piano sheet music identification task.

Most previous works on sheet music retrieval come from the literature on finding correspondences between audio and sheet music.  There are three general approaches to the cross-modal retrieval problem.  The first approach is to convert the sheet music into MIDI using optical music recognition (OMR), to compute chroma-like features on the MIDI, and then to compare the result to chroma features extracted from audio.  This approach has been applied to audio--sheet music synchronization \cite{DammFKMC08_MultimodalPresentationofMusic_ICMI}\cite{FremereyMC10_RepeatsJumps_ISMIR}\cite{KurthMFCC07_AutomatedSynchronization_ISMIR}\cite{ThomasFMC12_LinkingSheetMusicAudio_DagstuhlFU}, and it translates very naturally to retrieval applications like using a segment of sheet music to identify its corresponding audio recording \cite{FremereyMKC08_AutomaticMapping_ISMIR} or to retrieve the corresponding temporal passage from a specific audio recording \cite{FremereyCME09_SheetMusicID_ISMIR}.  The second approach is similar to the first, except that it replaces the full OMR with a mid-level feature representation based on the location of  noteheads relative to the staff lines \cite{izmirli2012bridging}\cite{yang2019midipassage}.  The third approach is to train a multimodal convolutional neural network to learn a latent feature space that directly encodes similarity between chunks of audio and sheet music snippets.  This approach has been applied to audio--sheet music alignment \cite{dorfer2016live}\cite{dorfer2017learning} and retrieval applications like using a snippet of audio to retrieve its corresponding sheet music snippet and vice versa \cite{dorfer2017learning}\cite{dorfer2018tismir}\cite{dorfer2018end}\cite{balke2019learning}.  See \cite{mueller2019cross} for an overview of work on cross-modal retrieval of music data.  Also, we note that a recent work \cite{waloschek2019identification} has proposed a neural network-based approach for finding corresponding measures between two different sheet music versions of a piece.

This current study differs from previous work in two ways.  First, we study the sheet music retrieval problem at a much larger scale.  Previous works have studied sheet music retrieval using searchable databases containing hundreds of sheet music scores or a few thousand short snippets of sheet music.  In contrast, we perform experiments using all solo piano sheet music scores in the entire International Library Music Score Project (IMSLP)\footnote{\url{https://imslp.org}} as a searchable database.  We believe that this is several orders of magnitude larger than any previous study on sheet music retrieval.  Second, we focus on queries that are cell phone images of sheet music.  Previous works have primarily focused on synthetic sheet music, scanned sheet music, and audio recordings as input queries.  While there have been a handful of works that study OMR on cell phone pictures of sheet music \cite{calvo2018camera}\cite{bui2014staff}\cite{vo2016mrf}\cite{blanes2017camera}\cite{vo2014distorted}, this area of study is still in its infancy.  Even though using cell phone pictures arguably makes the task much more challenging due to the additional sources of noise and distortion, we believe that this change leads to a much more practical, user-facing application.

Our approach to the piano sheet music identification task is to combine a recently proposed bootleg score feature representation with a novel hashing scheme.  The bootleg score feature was originally proposed for a MIDI--sheet music alignment task \cite{yang2019midipassage}.  A recent work has explored using the bootleg score features in a hashing framework for de-anonymizing files in the Lakh MIDI dataset by finding matches in sheet music data \cite{tsai2020towards}.  We will show that this previously proposed fingerprinting approach is far too slow for our current scenario.  Because our task is a real-time application, latency is an extremely important factor (unlike \cite{tsai2020towards}, which is an offline task).  In this work, we impose a hard constraint that our system must have an average runtime of $1$ second or less.  We diagnose the reason why this previously proposed fingerprinting scheme is slow, and we develop a novel fingerprinting scheme that is able to achieve our stringent runtime constraint.

This paper has three main contributions.  First, we propose a novel hashing scheme called dynamic n-gram fingerprinting.  This approach dynamically constructs n-gram fingerprints of variable length in order to ensure that each fingerprint is discriminative enough to warrant a table lookup.  Second, we present empirical validation of our proposed method on a very large-scale retrieval task.  We perform experiments using all solo piano scores in IMSLP as a searchable database.  We show that dynamic n-gram fingerprinting achieves both higher retrieval accuracy and significantly lower runtimes than a previously proposed approach.  Our best system achieves a mean reciprocal rank of $0.85$ and has an average runtime of $0.98$ seconds per query.  Third, as a byproduct of this project, we release the precomputed bootleg score features on all piano scores in IMSLP.\footnote{Code for the paper can be found at \url{https://github.com/HMC-MIR/SheetMusicID}, and the precomputed bootleg score features can be found at \url{https://github.com/HMC-MIR/piano_bootleg_scores}.}  Because this task required a tremendous amount of time and computation involving the use of a supercomputing infrastructure, we release the features as a standalone repository in the hopes that it will be useful in a variety of other MIR-related tasks.


\section{System Description}
\label{sec:systemDescr}

Figure \ref{fig:systemOverview} shows the architecture of our proposed system.  We will describe the system in two parts: constructing the database and performing a search at runtime.

\subsection{Database Construction}
\label{subsec:db}

\begin{figure}
	\centerline{
		\includegraphics[width=\columnwidth]{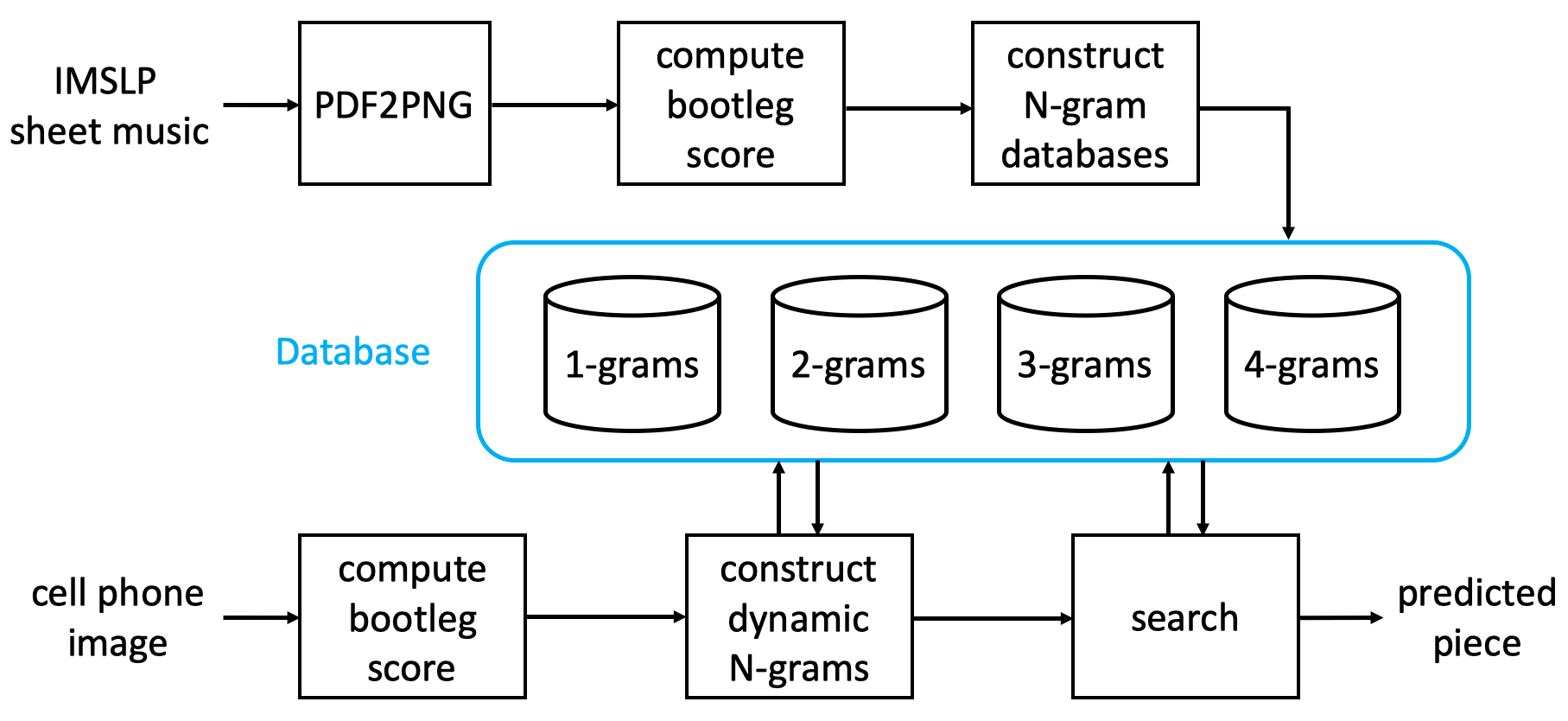}}
	\caption{Overview of proposed system.}
	\label{fig:systemOverview}
\end{figure}

Our first goal is to construct a database which will enable us to perform searches very efficiently.  The process of constructing this database consists of three steps, as shown in the upper half of Figure \ref{fig:systemOverview}.  These three steps will be described in the next three paragraphs.

The first step is to convert each sheet music PDF into a sequence of PNG images.  We decode the PDF into PNG images at 300 dpi, and then resize each image to have a width of 2550 pixels while preserving the aspect ratio.  Because there is an extremely large range of image sizes in the IMSLP dataset, we resize the images to ensure that they are within a range that the bootleg score feature computation was designed for.

\begin{figure}
	\centerline{
		\includegraphics[width=\columnwidth]{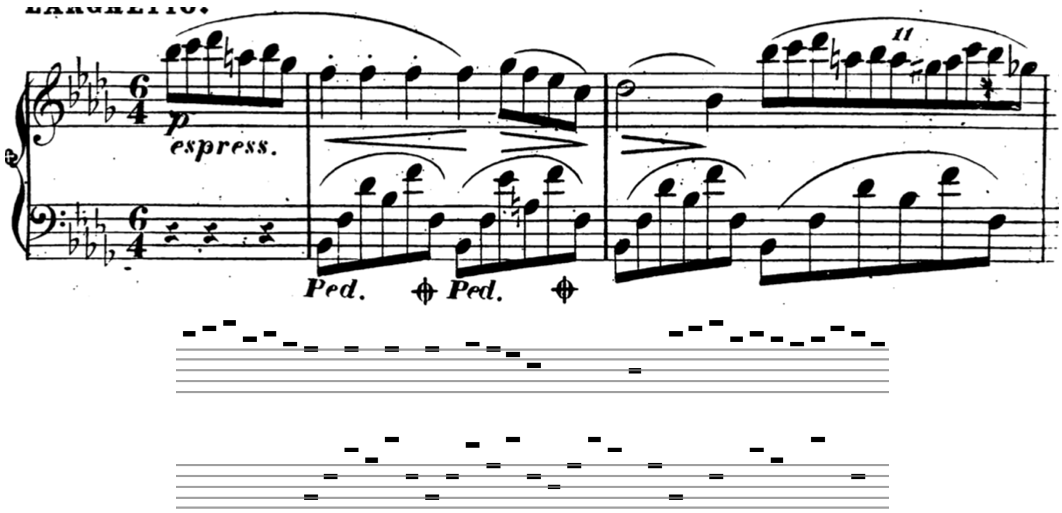}}
	\caption{A short section of sheet music and its corresponding bootleg score.}
	\label{fig:exampleBootlegScore}
\end{figure}

The second step is to compute a bootleg score for each page.  The bootleg score is a recently proposed feature representation of piano sheet music that encodes the position of filled noteheads relative to staff lines \cite{yang2019midipassage}.  The bootleg score representation itself is a $62 \times N$ binary matrix, where $62$ indicates the total number of possible staff line positions in both the left and right hands, and where $N$ indicates the total estimated number of simultaneous note events.  Figure \ref{fig:exampleBootlegScore} shows a short section of sheet music and its corresponding bootleg score representation.  Note that this representation discards a significant amount of information like duration, key signature, accidentals, octave markings, and clef changes, and it simply ignores non-filled noteheads (e.g. half or whole notes).  Nonetheless, it has been shown to be effective in aligning sheet music and MIDI, and we hypothesize that it may also be used effectively for large-scale retrieval.  The main benefit of using the bootleg score representation over a full OMR pipeline is processing time.  Because OMR is typically cast as an offline task, the best-performing systems require a significant amount of computation.\footnote{For example, in a recent survey on state-of-the-art music object detectors \cite{pacha2018baseline}, the best performing system required 40-80 seconds to process each image using a GPU.}  In contrast, the bootleg score can be computed on a high-resolution image in less than $1$ second using a CPU.  By focusing exclusively on simple geometrical shapes like circles (filled noteheads) and lines (staff lines and bar lines), it can detect objects robustly and efficiently using classical computer vision techniques.

The third step is to construct the n-gram databases.  The concept of an n-gram is adopted from the language modeling literature, where the likelihood of a sequence of $N$ consecutive words is estimated based on the frequency of its occurrence in a large set of data.  Here, we treat each bootleg score column as a word and consider $N$ consecutive words as a single fingerprint.  We generate four separate n-gram databases for $N=1,2,3,4$.  Each n-gram database is constructed in the following manner.  First, we concatenate the bootleg score features from all pages into a single, global bootleg score for each PDF.  Second, we represent each bootleg score column as a single 64-bit integer.  This allows us to represent the bootleg score very compactly as a sequence of integers.  Third, we consider every n-gram in the sequence as a fingerprint.  For example, if a bootleg score is given by a sequence of 64-bit integers $x_1, x_2, x_3, \cdots$, then the set of 3-gram fingerprints for this bootleg score is given by $(x_1, x_2, x_3), (x_2, x_3, x_4), (x_3, x_4, x_5), \cdots$.  Fourth, we store the location information for each fingerprint in a reverse index.  For each n-gram database, the hash key is a $(64 \cdot N)$-bit fingerprint, and the reverse index stores a list of $(\mathit{id}, \mathit{offset})$ tuples for all occurrences of that fingerprint in the database, where $\mathit{id}$ is a unique identifier for the PDF and $\mathit{offset}$ specifies the offset in the bootleg score.

\subsection{Search}
\label{subsec:search}

At runtime, our goal is to identify the piece of music showing in a cell phone image query.  This process consists of three steps, as shown in the bottom half of Figure \ref{fig:systemOverview}.

The first step is to compute a bootleg score on the cell phone image.  This is done using the same feature extraction as in the database construction phase.  Note that the inputs in the offline and online phases are very different: whereas the IMSLP data is primarily digital scans of physical sheet music, the queries are cell phone images of physical sheet music.  This introduces a lot of noise due to variable lighting conditions, zoom, camera angle, cropping, blur, unwanted objects outside the boundaries of the page, etc.  The only reason we can get away with using the same feature extractor in these two very different scenarios is that the bootleg score feature extraction has no trainable weights and only a small set of hyperparameters.  This makes it less likely to highly overfit to a set of data.  It is also worth pointing out that the bootleg score feature extraction was originally designed to handle the challenging case of cell phone images of sheet music, so we surmise that it will handle the easier case of scanned sheet music reasonably well.

The second step is to construct a sequence of dynamic n-gram fingerprints.  We will explain and motivate the use of dynamic n-grams by describing our initial attempts to solve the problem, the issues with these earlier approaches, and how the dynamic n-gram addresses these issues.

Our initial attempt was to consider each column of the bootleg score as a fingerprint.  This is equivalent to a 1-gram in the terminology used in this paper.  This approach was proposed in a recent work that attempts to de-anonymize files in the Lakh MIDI dataset by finding matches in a set of known sheet music data \cite{tsai2020towards}.  When we implemented this approach, we found that the retrieval accuracy was good, but that the system was far too slow.  This is an acceptable solution in \cite{tsai2020towards} because the task is offline, but it is an unacceptable solution in our current application because we have a very stringent runtime constraint.  Upon further analysis, we found that the frequency distribution of fingerprints was highly peaked and thus ill-suited for hashing.  In other words, there was a small set of fingerprints that occurred very frequently in the database.  These fingerprints tended to be bootleg score columns containing a single note event.  Because this occurs so frequently in piano sheet music, it forces the system to process an extremely large number of spurious fingerprints at runtime, which significantly slows down the system.

Our second attempt was to use an n-gram fingerprint to address this issue.  This introduces a tradeoff.  On the one hand, as we increase $N$ the fingerprint becomes more discriminative, which leads to fewer matches in the database and faster runtime.  On the other hand, increasing $N$ increases the likelihood that the fingerprint is erroneous, since the entire fingerprint is wrong if even one of its elements has an error.  If we roughly model each n-gram as $N$, independent Bernoulli random variables, then the probability that the entire n-gram is correct decreases exponentially in $N$.  Given this tradeoff, one very reasonable approach is to try different values of $N$ and to select the value that yields the best performance.

The dynamic n-gram gets the best of both worlds.  If a single bootleg word $x_i$ (i.e. a bootleg score column converted to a 64-bit integer) is very distinctive, then we simply do a table lookup in the 1-gram database.  In this case, it would not benefit us to do a 5-gram lookup if the first element only occurs a few times in the whole database.  If, however, the bootleg word is very common, then we prefer not to do a table lookup on the 1-gram database because this would require us to process a large number of spurious fingerprints.  In this case, we bump the 1-gram up to a 2-gram and repeat the process.  If the 2-gram fingerprint $(x_i, x_{i+1})$ is distinctive, then we do a table lookup on the 2-gram database.  If $(x_i, x_{i+1})$ is not distinctive, then we bump the 2-gram up to a 3-gram $(x_i, x_{i+1}, x_{i+2})$.  We repeat this process until the fingerprint is distinctive enough to warrant doing a table lookup (up to $N=4$).  As an example, given a sequence of bootleg words $x_1, x_2, x_3, \cdots$, one possible dynamic n-gram sequence would be $(x_1), (x_2, x_3), (x_3), (x_4, x_5, x_6), \cdots$.  Note that there is only one hyperparameter $\gamma$ that specifies the maximum number of fingerprint matches we are willing to process for every table lookup.

The third step is to search the database using the histogram of offsets method.  The histogram of offsets was proposed in \cite{wang2003industrial} as a way to efficiently search a very large database.  It is based on the observation that the true match in the database will yield a sequence of matching fingerprints at an approximately constant relative offset.  For example, if a query bootleg word sequence $x_1, x_2, x_3, \cdots$ matches a reference sequence $\widetilde{x}_i, \widetilde{x}_{i+1}, \widetilde{x}_{i+2}, \cdots$, then the matching fingerprints would all have a relative offset of $i-1$.  If we compute a histogram of relative offsets for all matching fingerprints with that item in the database, we would see a large spike in the histogram at the bin corresponding to an offset of $i-1$.  We can therefore compute a similarity score by constructing a histogram of offsets for matching fingerprints, and then calculate the maximum bin count in the histogram.  Once we have calculated a match score for every PDF in the database in this way, we group the PDFs by piece.  The \textit{piece} match score is calculated as the maximum score of any of its constituent PDFs.  Finally, we sort the pieces in the database by their match scores.  This yields our final predicted ranked list of pieces.

\section{Experimental Setup}
\label{sec:setup}

In this section, we describe the data and metrics used to evaluate our proposed system.

The cell phone image queries were taken from the Sheet MIDI Retrieval dataset \cite{tsai2020using}.  We include a description of the dataset here for completeness.  This data was originally created to study the task of aligning a cell phone picture of piano sheet music and its corresponding MIDI file.  It contains 2000 cell phone images of 200 different piano pieces across 25 well-known composers.  For each of the 200 pieces, a PDF from IMSLP was downloaded and printed onto physical paper.  Ten cell phone pictures were taken across the length of each piece in a variety of locations, lighting conditions, perspectives, and levels of zoom.  The pictures contain between 1 and 5 lines of music and were all taken in landscape orientation.  We use the same train-test split as the original paper: 400 cell phone images (corresponding to 40 pieces) were used for training, and 1600 cell phone images (corresponding to 160 pieces) were used for testing.  By using the same train-test split as the original paper, we ensure that the bootleg score feature extraction has not been tuned to the test data.

The database comes from IMSLP.  We first scraped the website and downloaded all PDF scores and associated metadata.\footnote{We downloaded the data over the span of several weeks in May 2018.}  We then filtered the data by instrumentation tag label in order to identify a list of solo piano pieces.  The resulting dataset contains 29,310 pieces and 31,384 PDFs and 374,758 individual images.  This is the dataset that we used to construct the database described in Section \ref{subsec:db}.  

We release the precomputed bootleg score features for all piano scores in the IMSLP dataset in a separate standalone repository.\footnote{\url{https://github.com/HMC-MIR/piano_bootleg_scores}}  We believe this is in itself a significant contribution to the MIR research community, given the amount of time, memory, and computation required to generate it.  For example, it took us over a month to scrape the IMSLP website and download all the scores in PDF format.  This set of PDFs was approximately 1.2 terabytes in size.  If the PDFs had been decoded into high-resolution images, the dataset would be in the tens of terabytes.  Because this was too large to store on disk, we decompressed each PDF to a set of high-resolution images, computed the bootleg score features, and then deleted the high-resolution images to conserve disk space.  We performed all feature computation on the NSF XSEDE supercomputing infrastructure \cite{xsede}.

We evaluate our system along two dimensions: retrieval accuracy and runtime.  Because our goal is to identify the matching \textit{piece} rather than just the exact same PDF,\footnote{In Section \ref{subsec:analysisSheetVersion}, we will test how well our system can identify a piece when only an alternate edition of the sheet music exists in the database.} there is always exactly one correct item in the database.  Accordingly, we use mean reciprocal rank (MRR) as our measure of retrieval accuracy.  The MRR is calculated as
\begin{equation*}
MRR = \frac{1}{N} \sum_{i=1}^{N} \frac{1}{R_i}
\end{equation*}
where $N=1600$ indicates the number of test queries and $R_i$ indicates the rank of the true matching item for the $i^{th}$ query.  In our task, $R_i$ can range between $1$ and $29310$, the total number of pieces in the database.  MRR ranges between $0$ and $1$, where $1$ indicates perfect performance.  We also measure the runtime required to process each test query.  The runtime includes all data pre-processing such as converting the JPG image to PNG format.  Note, however, that the runtimes do \textit{not} include the network latency that would be present in a real cell phone application.  All experiments are done on a single core of a 2.1 GHz Intel Xeon CPU.

\section{Results}
\label{[sec:results}

In this section, we present our experimental results on the piano sheet music identification task.

We are not aware of previous work that directly studies sheet music identification based on cell phone images.  As mentioned in Section \ref{sec:introduction}, there are works in the audio--sheet music alignment literature that have studied cross-modal sheet music retrieval.  These works could in principle serve as baseline comparisons.  However, all of the works we are aware of would not have been practical to evaluate on our task for one of two reasons.  First, some approaches do not scale to a large database.  For example, approaches that use subsequence DTW \cite{FremereyMKC08_AutomaticMapping_ISMIR}\cite{FremereyCME09_SheetMusicID_ISMIR}\cite{FremereyMC10_RepeatsJumps_ISMIR}\cite{yang2019midipassage} would have exorbitantly high runtimes on the IMSLP database.  Second, some approaches might have acceptable runtimes at test time, but would have required too much computation to construct the database.  For example, the sheet--audio alignment system in \cite{dorfer2018tismir} is 20 times slower than our proposed system and would have exceeded our computational budget on the XSEDE supercomputing infrastructure.  Any approaches that use OMR to convert the sheet music into MIDI format would likewise be too computationally expensive.

Nonetheless, we compare our proposed approach to nine other baseline systems.  The first four baselines are representative of the state-of-the-art in image retrieval in the computer vision community.  These systems were developed for the Oxford 5k \cite{philbin2007object} and Paris 6k \cite{philbin2008lost} benchmarks, where the goal is to identify a famous landmark in a query image given a database of known images.  All four systems are built on top of pretrained ImageNet classifiers like VGG \cite{simonyan2014very} and ResNet \cite{he2016deep}, but they differ in the method by which they convert model activations into a final feature representation.  The first baseline (MAC \cite{radenovic2016cnn}) takes the $K \times W \times H$ tensor of activations at the last convolutional layer and computes the maximum activation within each feature map.  This yields a fixed-size $K$-dimensional feature representation regardless of the image size.  The second baseline (SPoC \cite{babenko2015aggregating}) adopts a similar approach, but uses average pooling rather than max pooling.  The third baseline (GeM \cite{radenovic2018fine}) uses generalized mean pooling, which is a generalization of both average and max pooling where the type of pooling is specified by a single, trainable parameter.  The fourth baseline (R-MAC \cite{tolias2016particular}) applies max pooling over different regions of the image at various scales and combines the results through another pooling stage.  All four baselines also apply various forms of post-processing, such as dimensionality reduction through principal component analysis, whitening, and L2 normalization.  Given a query feature representation, similarity with database images is computed with a simple inner product.  In our experiments, we compute piece similarity as the maximum similarity with any page in any of the piece's constituent PDFs.  We evaluate the baseline systems with their provided pretrained models.\footnote{Training the baseline systems from scratch would require a large amount of labeled data (to retrain the ImageNet classifier) and would constitute a significant research project on its own.  In this work, we simply evaluate the baseline systems out-of-the-box using the provided pretrained models.}  The last five baselines are equivalent to our proposed system but using a fixed n-gram fingerprint for $N=1,2,3,4,5$.  The 1-gram system corresponds to the approach proposed in \cite{tsai2020towards}.

\begin{table}
	\begin{center}
		\begin{tabular}{|l|cc|cc|}
			\hline
			\multirow{2}{*}{System} &
			\multicolumn{2}{|c|}{MRR} & \multicolumn{2}{|c|}{Runtime}\\
			& cond 1 & cond 2 & avg & std \\
			\hline
			MAC \cite{radenovic2016cnn} & $.037$ & $.043$ & $1.17$s & $.12$s \\
			SPoC \cite{babenko2015aggregating} & $.003$ & $.004$ & $1.14$s & $.10$s \\
			GeM \cite{radenovic2018fine} & $.025$ & $.029$ & $1.18$s & $.11$s \\ 
			R-MAC \cite{tolias2016particular} & $.036$ & $.039$ & $.96$s & $.11$s \\
			\hline
			1-gram \cite{tsai2020towards} & $.709$ & $.659$ & $21.5$s & $12.5$s \\
			2-gram & $.845$ & $.784$ & $2.76$s & $1.11$s \\
			3-gram & $.808$ & $.767$ & $1.99$s & $.36$s \\
			4-gram & $.755$ & $.722$ & $1.12$s & $.25$s \\
			5-gram & $.688$ & $.668$ & $1.07$s & $.13$s \\
			\hline
			dynamic n-gram & $.853$ & $.812$ & $.98$s & $.12$s \\
			\hline
		\end{tabular}
	\end{center}
	\caption{System performance on the piano sheet music identification task.  Condition 1 is when the exact same PDF exists in the database.  Condition 2 is when only an alternate version of the sheet music is in the database.}
	\label{tab:results}
\end{table}

Table \ref{tab:results} compares the performance of all models.  The systems are presented in three groups: the image retrieval baselines (top), the fixed n-gram systems (middle), and the proposed dynamic n-gram system (bottom).  The second column (labeled ``cond 1") indicates the MRR on the test set, and the last two columns indicate the average runtime per query and corresponding standard deviation.  The column labeled ``cond 2" will be discussed in Section \ref{subsec:analysisSheetVersion}.

There are  three things to notice about these results.  First, the image retrieval baselines all perform very poorly.  The best-performing image retrieval system is MAC, which achieves an MRR of $.037$.  This is not a surprise, since these systems were not designed for working with sheet music images, but it does confirm that existing image retrieval systems do not work out-of-the-box on the sheet music identification task.  We do observe, however, that the systems achieve results significantly better than random guessing (approximately $.001$ MRR).  Second, the fixed n-gram systems show a tradeoff between retrieval accuracy and runtime.  As $N$ increases from 1 to 5, we see the average runtime decrease from $21.5$ seconds to $1.07$ seconds.  This reflects the fact that the fingerprint is becoming more and more discriminative, which leads to fewer and fewer matches in the database.  At the same time, we observe that the retrieval accuracy decreases from $.845$ to $.688$ as $N$ increases from 2 to 5.  This reflects the fact that more and more fingerprints are erroneous as fingerprint size increases.  The increase in MRR from $N=1$ to $N=2$ indicates that the 1-gram fingerprints are not sufficiently distinctive.  Third, the dynamic n-gram system achieves both the highest retrieval accuracy ($.853$) and the lowest average runtime ($0.98$ seconds).  This indicates that the design has achieved its intended goal: to avoid the tradeoff between retrieval accuracy and runtime, and to instead get the best of both worlds.

\section{Analysis}
\label{sec:analysis}

In this section, we conduct four different analyses to answer key questions of interest.

\subsection{Failure Modes}

The first question of interest is, ``What are the failure modes of the system?"  To answer this question, we identified the queries with poor reciprocal rank values and investigated the reasons for failure.  By far the biggest reason for poor performance was failure in the bootleg score feature computation.  Common mistakes included missed detection of non-filled noteheads or noteheads occurring in block chords, notehead detection false alarms arising from text and other musical symbols on the page, and staff line estimation errors.  Fixing these issues would require re-designing the bootleg feature computation.  Another (minor) reason for poor performance came from non-distinctive sections of music.  For example, when there are repetitive octaves or long sequences of alternating between two notes in only one hand, this can have a strong match with unrelated pieces of music.

\subsection{Effect of Sheet Music Version}
\label{subsec:analysisSheetVersion}

The second question of interest is, ``How well does the system handle different sheet music versions?"  To answer this question, we ran a separate set of experiments in which we remove the exact same PDF from the database.  This means that the system can only match against alternate versions of the sheet music.  Because some queries only had 1 sheet music version in IMSLP, this additional benchmark was run on a reduced subset of 930 test queries.

The results of this alternate benchmark are indicated in Table \ref{tab:results} as ``Condition 2."  We see that the MRR of the fixed n-gram systems has been reduced somewhere between $.02$ and $.06$, and the MRR of the dynamic n-gram system is reduced by $.04$.  This performance gap between condition 1 and condition 2 can be interpreted as the additional performance loss that is caused by variations in different sheet music editions.  While this is a nontrivial decrease in performance, the proposed system still has a robust overall retrieval accuracy ($.812$ MRR) when the exact same version is not in the database.

\begin{table}
	\begin{center}
		\begin{tabular}{|l|cc|cc|}
			\hline
			\multirow{2}{*}{System} &
			\multicolumn{2}{|c|}{MRR} & \multicolumn{2}{|c|}{Runtime}\\
			& cond 1 & cond 2 & avg & std \\
			\hline
			dyn n-gram (20k) & $.864$ & $.802$ & $1.67$s & $.18$s \\
			dyn n-gram (10k) & $.865$ & $.802$ & $1.53$s & $.16$s \\
			dyn n-gram (5k) & $.860$ & $.803$ & $1.21$s & $.15$s \\
			dyn n-gram (1k) & $.853$ & $.812$ & $.98$s & $.12$s \\			
			\hline
		\end{tabular}
	\end{center}
	\caption{Comparison of dynamic n-gram systems with various values of $\gamma$, which specifies the maximum number of fingerprint matches the system will process on a table lookup before bumping an $N$-gram to an $(N+1)$-gram.}
	\label{tab:effectGamma}
\end{table}

\subsection{Effect of $\gamma$}

The third question of interest is, ``How does system performance vary with $\gamma$?"  Recall that the dynamic n-gram approach has one hyperparameter $\gamma$ that specifies the maximum number of fingerprints we are willing to process for each table lookup.  We ran experiments with several values of $\gamma$ to determine its effect on system performance.

Table \ref{tab:effectGamma} shows system performance for $\gamma$ ranging from $1000$ to $20,000$.  As $\gamma$ decreases, we see a very slight decrease in retrieval accuracy ($.864$ to $.853$) and significant improvement in average runtime ($1.67$s to $.98$s).  In this case, we have a very nice tradeoff: for only a small sacrifice in retrieval accuracy, we can significantly speed up the system.  The dynamic n-gram results in Table \ref{tab:results} correspond to $\gamma = 1000$, which is the best system that meets our constraint of $1$ second average runtime per query.

\subsection{Fingerprint Distribution}

The fourth question of interest is, ``How well suited for hashing is the dynamic n-gram fingerprint distribution?"  As described in Section \ref{subsec:search}, we found that the 1-gram fingerprint proposed in \cite{tsai2020towards} had a frequency distribution that was very peaked and thus ill-suited for hashing.\footnote{Note that the ideal distribution for hashing is a uniform distribution.}  To see how well the dynamic n-gram approach addresses this issue, we compared its frequency distribution to the fixed n-gram approaches.

\begin{figure}
	\centerline{
		\includegraphics[width=\columnwidth]{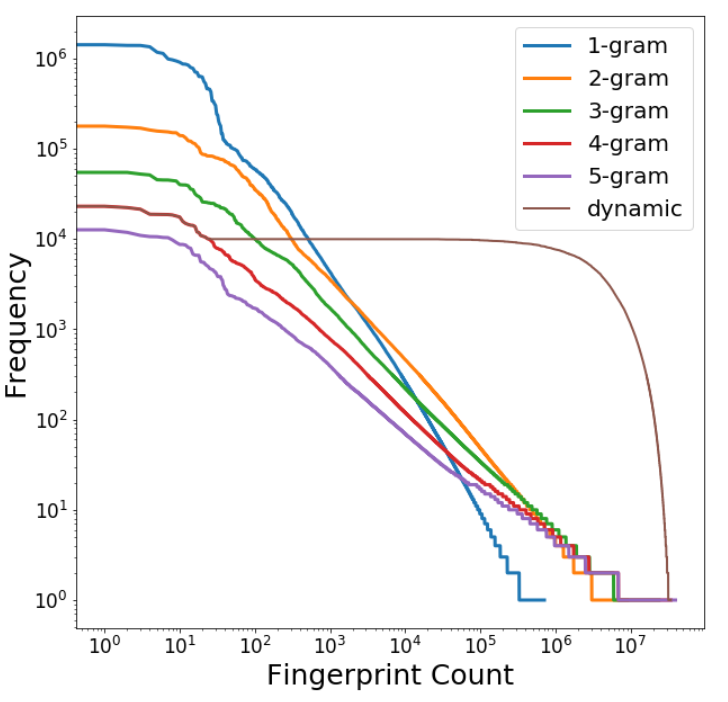}}
	\caption{Comparing the fingerprint frequency distributions of the fixed n-gram systems and dynamic n-gram with $\gamma=10,000$.  For each curve, the fingerprints have been ordered from most frequent (left) to least frequent (right).}
	\label{fig:fpDistributions}
\end{figure}

Figure \ref{fig:fpDistributions} shows this comparison.  Each curve shows the frequency of different fingerprint values, where the fingerprints have been sorted from most frequent (left) to least frequent (right).  Both axes are shown on a log scale in order to better visualize the wide dynamic range.  Note that all of the fixed n-gram distributions have approximately the same total number of fingerprints in the database, so their only difference is how many unique fingerprint values there are and how the fingerprints are distributed across these different values.  The curve for the dynamic n-gram system corresponds to $\gamma=10,000$.

Figure \ref{fig:fpDistributions} shows the same trends that we see in Table \ref{tab:results}.  The difference, however, is that Figure \ref{fig:fpDistributions} explains \textit{why} the results in Table \ref{tab:results} are the way they are.  For example, we observed earlier that the fixed n-gram systems exhibit a tradeoff between retrieval accuracy and runtime.  Figure \ref{fig:fpDistributions} explains this tradeoff from a hashing perspective.  As $N$ increases, the fixed n-gram distributions become less and less peaked, which translates to fewer fingerprint matches in the database and smaller runtimes.  At the same time, this manner of reducing the peak comes with an exponential explosion in the number of unique fingerprint values.  This means that the fingerprints will be less generalizable and more error-prone.  We also notice that the only fixed n-gram curves that intersect early on are the 1-gram and 2-gram curves.  This explains why the 2-gram system is unilaterally better than the 1-gram system: it has a less peaked distribution (smaller runtime) and it has a higher frequency of fingerprints than the 1-gram system for a large fraction of fingerprint values (better retrieval accuracy).  Finally, we observe that the dynamic n-gram system has a flatter distribution than any of the fixed n-gram systems across a wide range of its distribution.  This confirms that the dynamic n-gram approach is able to transform the distribution into one that is more well-suited for hashing.\footnote{Note that the curve for the dynamic n-gram system has about 30 fingerprints that occur more than $\gamma=10,000$ times.  This is because we do not include n-grams higher than $N=4$ in our database.}

\section{Conclusion}
\label{sec:conclusion}

We present a method for identifying a page of piano sheet music given a cell phone picture taken of it.  Our approach combines the bootleg score feature representation with a novel hashing scheme.  We show that a previously proposed 
fingerprinting approach yields reasonable retrieval accuracy, but is far too slow to be useful in a real-time application.  We propose a novel dynamic n-gram fingerprinting scheme, in which we dynamically construct n-grams of variable length in order to ensure robust hashing performance.  We perform experiments using all solo piano sheet music scores in the entire IMSLP dataset as a searchable database, and our system achieves a mean reciprocal rank of $.85$ with an average runtime of $.98$ seconds per query.  For future work, we would like to explore the task of identifying non-piano sheet music.

\section{Acknowledgments}
This work used the Extreme Science and Engineering Discovery Environment (XSEDE), which is supported by National Science Foundation grant number ACI-1548562.  Large-scale computations were performed with XSEDE Bridges at the Pittsburgh Supercomputing Center through allocation TG-IRI190019.

\bibliography{sheetMusicID}

\end{document}